\documentclass{article}
\usepackage{spconf,amsmath,graphicx,booktabs}

\usepackage{booktabs}
\usepackage{multirow}
\usepackage{microtype}

\def\fleurs{FLEURS }
\title{Self-supervised Adaptive Pre-training of Multilingual Speech Models for Language and Dialect Identification}
\name{Mohammed Maqsood Shaik  \hspace{0.6cm}  Dietrich Klakow \hspace{0.6cm} Badr M. Abdullah}      
\address{Language Science and Technology, Saarland University, Germany \\  }
\begin{document}
%
\maketitle
\begin{abstract}
Transformer-based, pre-trained speech models have shown striking performance when fine-tuned on various downstream tasks such as automatic speech recognition and spoken language identification (SLID). However, the problem of domain mismatch remains a challenge in this area, where the domain of the pre-training data might differ from that of the downstream labeled data used for fine-tuning. In multilingual tasks such as SLID, the pre-trained speech model may not support all the languages in the downstream task. To address this challenge, we propose self-supervised adaptive pre-training (SAPT) to adapt the pre-trained model to the target domain and languages of the downstream data. We apply SAPT to the XLSR-128 model and investigate the effectiveness of this approach for the SLID task. First, we demonstrate that SAPT improves XLSR’s performance on the FLEURS benchmark with substantial gains up to 40.1\% for under-represented languages. Second, we apply SAPT on four different datasets in a few-shot learning setting, showing that our approach improves the sample efficiency of XLSR during fine-tuning. Our experiments provide strong empirical evidence that continual adaptation via self-supervision improves downstream performance for multilingual speech models.

\end{abstract}
\section{Introduction}
\label{sec:intro}

Self-supervision has proven to be an effective approach for learning universal representations of spoken language directly from raw audio, obviating the need for explicit labels or transcriptions  (refer to \cite{9893562} for a thorough overview). 
Transformer-based self-supervised models such as wav2vec 2.0 \cite{baevski2020wav2vec} and HuBERT \cite{hsu2021hubert} excel on a spectrum of speech processing tasks.
These models can be fine-tuned with labeled data for tasks such as automatic speech recognition and speaker identification. 
Fine-tuning these models have shown striking performance especially when limited labeled are available data for downstream tasks \cite{chen2022wavlm, baevski2020wav2vec, hsu2021hubert}.

However, an open challenge in this area of research is the mismatch between the unlabeled data used for pre-training and the labeled data for fine-tuning, often referred to as the \textbf{domain shift} \cite{domain_shift} problem in machine learning research.
In speech modality, the term domain is used to denote a characterization of the input samples in terms of different factors \cite{hsu21_interspeech, sanabria2022measuring, zuluaga2023does}: (1) acoustic conditions (e.g., clean vs. noisy speech, etc.), (2) linguistic variation (e.g., non-native or dialectical speech), or  (3) genre (read speech vs. spontaneous speech). 
To address the domain mismatch, strategies such as self-supervised continual pre-training has been proposed and found to be effective for text \cite{gururangan2020don} and vision \cite{liu2023unsupervised}. 
In this approach, unlabeled samples from the target domain or downstream task are used to adapt the pre-trained model before the final fine-tuning step.

\begin{figure}[t]
  \centering
  \includegraphics[width=0.85\linewidth]{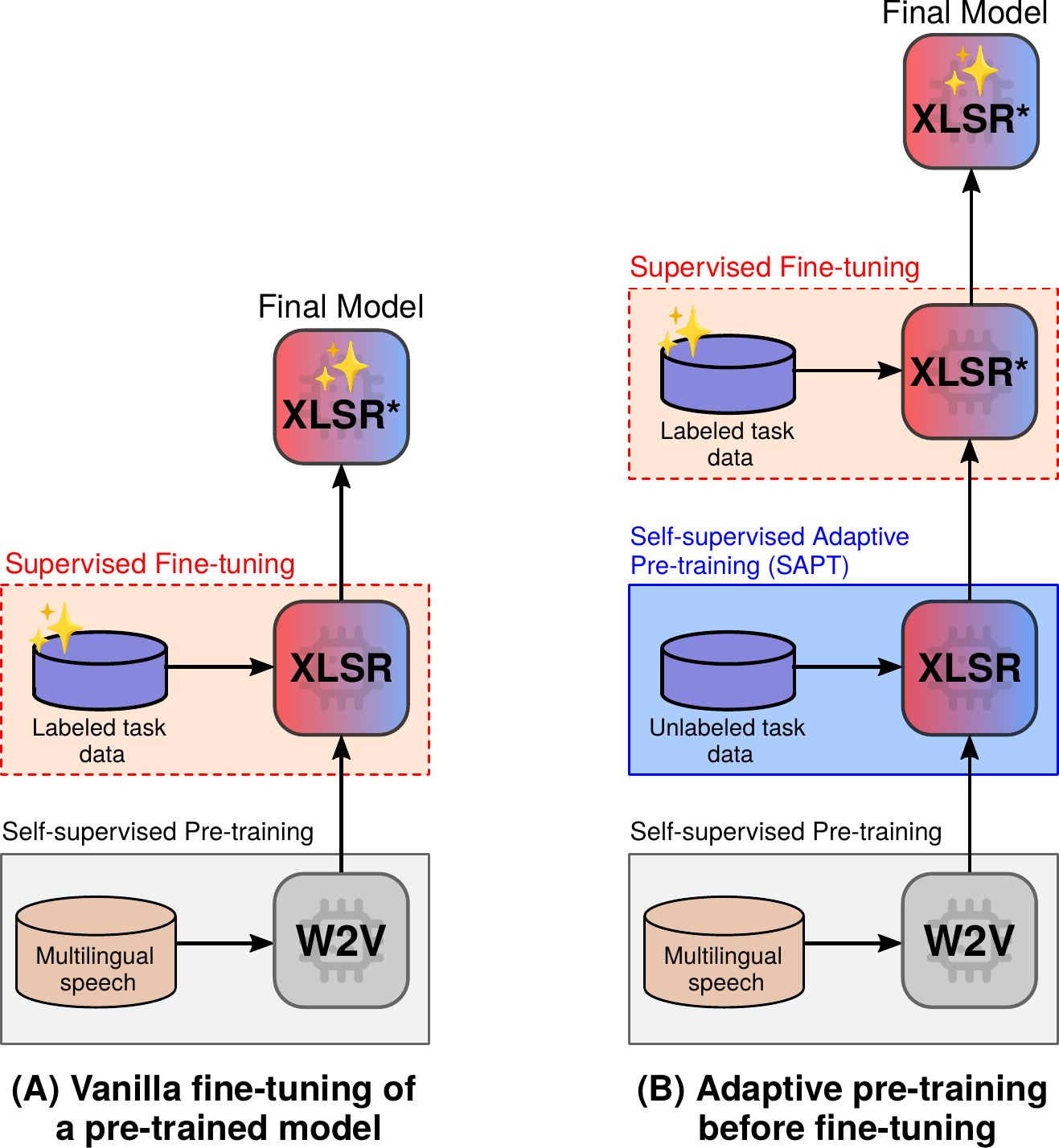}
  \caption{A visual illustration of two approaches for adapting a pre-trained model to a target task: (A) vanilla fine-tuning, and (B) self-supervised adaptive pre-training (SAPT) on unlabeled samples from the target dataset before fine-tuning, applied to XLSR speech model. In this paper, we propose SAPT of multilingual speech models for the task of spoken language identification (SLID). }
  \label{fig:fig_1}
\end{figure}

Another key feature of pre-trained speech models is that they can also be pre-trained using multilingual speech corpora \cite{conneau2020unsupervised, DBLP:conf/interspeech/BabuWTLXGSPSPBC22}.
This setup can introduce a unique set of domain mismatch challenges. 
A language might be part of the pre-training data but differ in genre from the target domain, or perhaps, not be included in the multilingual model at all.
As a result, a domain mismatch usually occurs due to a combination of the aforementioned acoustic and linguistic factors. 
Hence, the effectiveness of continual pre-training strategies on multilingual self-supervised models remains an open question.

In this paper, we extend self-supervised adaptive pre-training (SAPT) to a multilingual setting by continually pre-training the XLS-R-128 model \cite{DBLP:conf/interspeech/BabuWTLXGSPSPBC22} on unlabeled samples  using datasets that are not seen during pre-training. 
We focus on the task of spoken language identification (SLID) for our experiments. 
Concretely, we first show that SAPT yields substantial improvements on the common FLEURS benchmark with up to +40.1 relative gain in accuracy (\S\ref{sec:FLEURS}).  Then, we demonstrate that SAPT is an effective approach in a few-shot setting on four different SLID datasets for both high-resource and low-resource language varieties (\S\ref{sec:FEW-SHOT}).

\section{CONTINUAL SELF-SUPERVISED PRE-TRAINING}
\label{sec:formulation}
In the section, we present a formal mathematical description of our proposed approach and how it is different from the conventional pre-training then fine-tuning paradigm. \vspace{0.5cm}

\noindent
\textbf{Self-supervised pre-training} \hspace{0.25cm} Consider a large, unlabeled pre-training speech corpus $\mathcal{D}_{0}$ with an individual input sample denoted as $\mathbf{s}$. 
In the initial phase of pre-training, model parameters $\boldsymbol{\theta}$ are randomly initialized.
The pre-training objective can be defined as

\begin{equation}
J_{0}(\boldsymbol{\theta}) = \sum_{\mathbf{s} \in \mathcal{D}_{0}} \mathcal{L}_{SS}(\mathbf{s}, m(\mathbf{s}))
\end{equation}
Here, $ \mathcal{L}_{SS}(.)$ is the self-supervised loss and $m(.)$ is a function that constructs a supervision signal from the input sample itself. 
In our case, this is realized with a masked language modeling (MLM) task. 
The optimization during pre-training involves minimizing the pre-training loss such that
\begin{equation}
   \boldsymbol{\theta}_0 =  \arg\min_{\boldsymbol{\theta}} J_{0}(\boldsymbol{\theta})
\end{equation}

\noindent
\textbf{Supervised fine-tuning} \hspace{0.25cm} In this step, the pre-trained model is adapted to the target task using a  dataset $\mathcal{D}_{\mathcal{T}}$ consisting of labelled samples 
$(\mathbf{x}, \mathbf{y})$. 
The fine-tuning loss for a task-specific objective $\mathcal{L}_{{\mathcal{T}}}$ is defined as
\begin{equation}
\label{eq:FT}
J_{\mathcal{T}}(\boldsymbol{\theta}) = \sum_{(\mathbf{x}, \mathbf{y}) \in \mathcal{D}_{\mathcal{T}}} \mathcal{L}_{\mathcal{T}}(\mathbf{x}, \mathbf{y}, \boldsymbol{\theta}_0)
\end{equation}
Here, $\mathcal{L}_{\mathcal{T}}$ is a supervised loss for the downstream task such as cross-entropy for classification.
The optimization during fine-tuning involves minimizing the fine-tuning loss such that 
\begin{equation}
\label{eq:FT_opt}
   \boldsymbol{\theta}_\mathcal{T} =  \arg\min_{\boldsymbol{\theta}} J_{\mathcal{T}}(\boldsymbol{\theta})
\end{equation}

\noindent
\textbf{Self-supervised adaptive pre-training (SAPT)} \hspace{0.25cm}  Instead of directly fine-tuning the pre-trained model on the downstream task data, SAPT first adapts the model by continuing the pre-training objective on unlabeled samples from the target domain of the task data.  
The SAPT objective can be described as
\begin{equation}
J_{\text{SAPT}}(\boldsymbol{\theta)} = \sum_{\mathbf{x} \in \mathcal{D}_{\mathcal{T}}} \mathcal{L}_{SS}(\mathbf{x}, m(\mathbf{x}), \boldsymbol{\theta}_0)
\end{equation}
And we seek the parameters that minimize
\begin{equation}
   \boldsymbol{\theta}_{\text{SAPT}} =  \arg\min_{\boldsymbol{\theta}} J_{\text{SAPT}}(\boldsymbol{\theta})
\end{equation}
It is worth point out that the self-supervised pre-training objective remains unchanged, but the input samples are different and the parameters of the model during SAPT are the those of the already pre-trained model on $\mathcal{D}_{0}$. After the SAPT step, the model can be fine-tuned in a supervised setup using the labeled dataset $\mathcal{D}_\mathcal{T}$ as described in equations~(\ref{eq:FT}) and~(\ref{eq:FT_opt}), but the model parameters  in eq.~(\ref{eq:FT}) $\boldsymbol{\theta}_\text{SAPT}$ are instead of $\boldsymbol{\theta}_0$.

\section{Pre-trained Model: XLSR-128}
\label{sec:model}
We experiment with the multilingual XLSR-128 speech model  \cite{DBLP:conf/interspeech/BabuWTLXGSPSPBC22} since the main goal of our work is to investigate how self-supervision with downstream datasets affects final fine-tuning performance. 
The XLSR-128 model stands out for having effective and adaptable cross-lingual speech representations that were developed through pre-training the model on a large and diverse multilingual dataset.

\section{Experiment 1: The FLEURS Benchmark} 
\label{sec:FLEURS}
In this experiment, we apply the SAPT method for  language identification on the common \fleurs benchmark.

\begin{table*}[t]
\centering
\begin{tabular}{@{}ll|lcccccccc@{}}
\toprule
\textbf{} &  & \begin{tabular}[c]{@{}l@{}}geographic group\end{tabular} & WE & EE & CMN & SSA & SA & SEA & CJK & Avg. \\ \cmidrule(l){3-11} 
Model & size & Num of languages & 25 & 16 & 12 & 20 & 14 & 11 & 4 & 102 \\ \midrule
 w2v-bert-51 \cite{chung2021w2v}& 0.6B & & 85.3 & 78.4 & 72.9 & 59.1 & 52.0 & 65.7 & 89.7 & 71.4 \\
 mSLAM \cite{bapna2022mslam}& 0.6B & & 84.6 & 81.3 & 75.9 & 62.2 & 51.7 & 73.4 & 87.8 & 73.3 \\ 
XLSR \cite{DBLP:conf/interspeech/BabuWTLXGSPSPBC22}& 0.3B & & 79.2 & 93.7 & 93.6 & 67.2 & 75.1 & 81.5 & 99.8 & 84.3 \\
 XLSR + SAPT  (ours) & 0.3B & & \textbf{87.1} & 93.7 & \textbf{95.2} & \textbf{94.2} & \textbf{90.9} & \textbf{95.1} & \textbf{99.9} & \textbf{93.7} \\ \midrule
Gain w/ SAPT  ($\Delta$\%) &  & & \textbf{10.0} & -0.0 & \textbf{1.69} & \textbf{40.1} & \textbf{21.0} & \textbf{16.7} &  \textbf{0.1} &  \textbf{11.2} \\
\bottomrule
\end{tabular}
\caption{SLID performance on the FLEURS benchmark measured by accuracy (\%), categorized by geographical groups.}
\label{tab:fleurs}
\end{table*}

\subsection{Experimental Data and Setup}
 The \fleurs benchmark \cite{conneau2023fleurs} is a multilingual dataset for spoken language identification and translation, which is a spoken extension of the text-based FLORES benchmark  for Machine Translation \cite{goyal-etal-2022-flores}. 
 FLEURS comprises a collection of $\sim$2009 sentences per language extracted from the FLORES multi-way parallel evaluation set, spanning an 102 languages. 
 The languages are organized into seven geographic groups as: Western European (WE), Eastern Europe (EE), Central-Asia/MiddleEast/North-Africa (CMN), Sub-Saharan Africa (SSA), South Asia (SA), South-East Asia (SEA) and Chinese, Japanese, and
Korean (CJK) languages.
Furthermore, the data has three portions of 1109/400/500 utterances for the train/dev/test splits, respectively. 
Note that only the training portion of this dataset is used for SAPT, thus preventing any leakage from the final test set to the model. 
On average, the collected dataset contains about 2.3 utterances for each sentence.
Following prior experimental design in \cite{conneau2023fleurs}, we train a separate XLSR-128 model for each geographic group in two different settings: (1) vanilla fine-tuning, and (2) applying SAPT then fine-tuning.

\subsection{Experimental Results}
Table~\ref{tab:fleurs} shows the results of this experiment. 
We compare the performance of SAPT to the baselines established in \cite{conneau2023fleurs} which compares the performance of w2v-bert-51 \cite{chung2021w2v} and mSLAM \cite{bapna2022mslam} for this benchmark. 
We observe that vanilla fine-tuning of the XLSR-128 model outperforms the these strong baselines across most geographic groups, despite having less parameters. 
This improved performance can be attributed to pre-training on a larger number of languages, since XLSR was pre-trained on 128 languages compared to  51 languages in w2v-bert-51 (0.6B) and 101 languages in mSLAM (0.6B). 
The only exception to this trend is the case 
 WE (Western European) languages. 
 Since these languages are usually high-resource languages, this could be an effect of the so-called ``the curse of mulitlinguality'', where per-language performance degrades as more languages are included in the model  \cite{pfeiffer-etal-2022-lifting}. 

Next, we compare the performance of vanilla fine-tuning XLSR compared to our proposed SAPT approach. 
We observed that SAPT improves the performance for all geographic groups, increasing the macro-average accuracy from 84.3\% to 93.7\% and achieving a relative improvement of 11.2\%. 
Moreover, continual adaptive pre-training leads to substantial improvements for geographic groups that are low-resource and under-represented in the pre-training data. 
For example, the accuracy for the Sub-Saharan Africa (SSA) and South Asia (SA) groups improve by 40.1\% and 21.0\%, respectively. 
The only language group that does not benefit from SAPT before fine-tuning is the Eastern Europe (EE) group, yet the performance does not degrade compared to vanilla fine-tuning.
It is worth pointing out that the EE group is well-represented in the pre-trained XLSR model with $\sim$136.5k hours of speech, compared to $\sim$0.63k and  $\sim$0.91k for the SSA and SA group, respectively\footnote{these numbers correspond to the amount of pre-training data in the original  mixture of the languages during pre-training the wav2vec 2.0 architecture to produce the multilingual XLSR-128 model.}.
In summary, this experiment demonstrates that SAPT is an effective approach for adapting the speech multilingual model to a target downstream dataset, with substantial improvement for languages that are under-represented during pre-training.

\section{Experiment 2: Few-shot Learning}
 \label{sec:FEW-SHOT}
In this experiment, we apply the SAPT method for language identification on  four different diverse datasets that range from high-resource to low-resource languages and varieties.
The focus of this experiment is investigate the effectiveness of SAPT in a few-shot setting.

\begin{figure*}[t]
  \centering
  \includegraphics[width=0.99\linewidth]{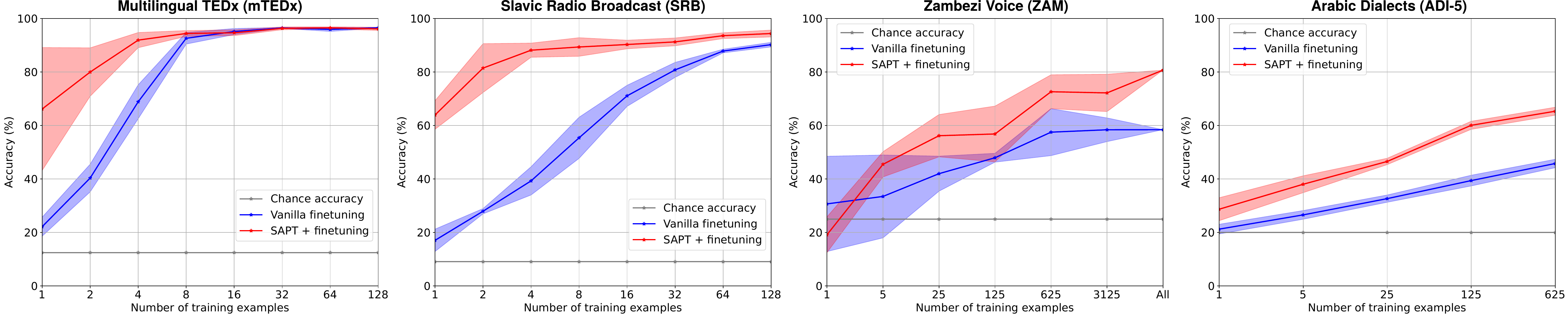}
  \caption{SLID performance in the few-shot setting by accuracy (\%) for four different multilingual datasets: vanilla fine-tuning (blue) vs. SAPT + fine-tuning (red).}
  \label{fig:prob_acc}
\end{figure*}

\subsection{Experimental Data}
\noindent
\textbf{Multilingual TEDx Talks (mTEDx)} \hspace{0.25cm}
A multilingual corpus that was collected from online TED Talks \cite{salesky2021mtedx}. 
This dataset contains 765 hours of audio in eight different languages: Arabic, German, Spanish, French, Portuguese, Italian, Russian, and Greek.
Even though these languages were encountered during the pre-training phase of the model XLSR-128, the domain of TED talks differs from pre-training data due to linguistic and acoustic factors. 

\vspace{0.25cm}

\noindent
\textbf{Slavic Radio Broadcast (SRB)} \hspace{0.25cm}
A large collection of Slavic recordings were collected by harvesting online radio broadcasts \cite{mateju2018using} and used in prior research for SLID \cite{abdullah2020rediscovering}. 
The dataset contains recordings for 11 Slavic languages: Czech, Russian, Serbian, Belarusian, Slovakian, Slovenian, Macedonian, Bulgarian, Polish, Ukrainian, and Croatian. 
This dataset is diverse in terms of recording environment with occasional background music and noise, which makes it ideal to study the effect of domain shift.
Also, the languages in this dataset are closely related, which makes it more difficult compared to the mTEDx dataset. 

\vspace{0.25cm}

\noindent
\textbf{Zambezi Voice (ZAM)} \hspace{0.25cm} 
An open-source  multilingual resource for the native languages of Zambia that includes recordings for five languages: Bemba, Nyanja, Tonga, Lozi, and Lunda \cite{sikasote23_interspeech} . 
This dataset consists of around 80 hours of labeled data and encompasses recorded read speech captured in uncontrolled environments, which might include instances of background noise.
Moreover, none of the languages were used during pre-training XLSR-128 model, therefore applying our proposed SAPT approach corresponds to language adaptive pre-training in addition to task adaptive pre-training. 

\vspace{0.25cm}

\noindent
\textbf{Arabic Dialects (ADI-5)} \hspace{0.25cm}
A collection of TV broadcast speech consisting of $\sim$60 hours was collected from YouTube for four different Arabic language varieties: Egyptian, Levantine, North African, Gulf, as well as Modern Standard Arabic (MSA) \cite{mgb3}.
Except for MSA, none of the arabic dialects in this dataset was used during pre-training XLSR. 
Therefore, this dataset is ideal for our study to investigate whether our approach can be extended to non-standard varieties that are not usually supported by large speech and language models.

\vspace{0.25cm}

\subsection{Experimental Results}
Fig.~\ref{fig:prob_acc} shows the results of this experiment across the four datasets.
The number of training samples was chosen based on the difficulty of each dataset.
To account for the variance in the case of few-shot learning when fine-tuning on a small number of sample, we fine-tune five runs with different random seeds for each setting. 
Also, each run is trained on a different sample of training instances. 
Similar to our first experiment with the FLEURS benchmark, only the training split of each dataset is used for SAPT. 
Therefore, samples from validation and test sets are never seen by the model before running the final evaluation.  
We observe that SAPT consistently leads to accuracy gains compared to vanilla fine-tuning.  
For the two datasets that mostly consists of high-resource languages and well-represented in the XLSR model (i.e., mTEDx and SRB), one can see that SAPT leads to better sample efficiency compared to vanilla fine-tuning.   
For mTEDX, SAPT yields substantial gains in the range 1-4 samples, while for SRB the gain is observed even for a wider range, that is, 1-64 samples.
Note that the mTEDx datasets contains languages that are linguistically distant and phonetically dissimilar, therefore is not surprising that the model reaches high accuracy given only a small number of samples.
On the other hand, the SRB dataset consists of the related Slavic languages that are very similar, therefore this dataset is more difficult for language identification compared to the mTEDx dataset, despite the fact that Slavic languages are well-represented in XLSR.

Regarding the datasets ZAM and ADI-5, which consists of languages that are not seen during the pre-training phase of XLSR except MSA, we observe that SAPT leads to improvement despite the fact that the performance gain is not as high as in the mTEDx and  SRB datasets. 
For the ZAM dataset, SAPT yields an improvement across different numbers of training samples including the case where all the samples  are used for fine-tuning whereby accuracy has notably risen from  from 58.42\% to 80.7\%, in comparison to the performance achieved through vanilla fine-tuning. 
This considerable increase in accuracy clearly demonstrates that SAPT is an effective approach, benefiting not only a few-shot fine-tuning scenario but also delivering substantial improvements across all sample sizes during the fine-tuning process.

\section{Conclusion and Future Work}
\label{sec:conclusions}
In summary, our paper tackles the challenge of domain mismatch in pre-trained, transformer-based speech models, specifically in multilingual tasks such as spoken language identification (SLID). 
We have proposed self-supervised adaptive pre-training (SAPT) to align these models with the specialized needs of downstream tasks. 
When applied to the XLSR-128 model, SAPT yielded notable performance gains, especially for under-represented languages, and improved sample efficiency in a few-shot learning setting.
Our results provide strong evidence that continual self-supervised adaptation is an effective strategy for improving the performance of multilingual speech models.
The main advantage of the SAPT approach is that it does not require any additional labeled data during fine-tuning and does not add any trainable parameters to the transformer architecture. 
For future work, the proposed SAPT approach can be extended for other speech tasks such as automatic speech recognition and speech translation.

\small{
\section*{Acknowledgement}  This research is funded by the Deutsche Forschungsgemeinschaft (DFG, German Research Foundation), Project ID 232722074 -- SFB 1102.}

\newpage

\bibliographystyle{IEEEbib}
\bibliography{icassp_paper}

\end{document}